\documentclass[11pt]{elsarticle}
\usepackage{lineno,hyperref}
\usepackage{optidef}
\usepackage{graphicx}
\usepackage{float}
\usepackage{graphicx}
\usepackage{subfig}
\usepackage{mathrsfs}
\usepackage{float}
\usepackage{amsmath}
\usepackage{amsfonts}
\usepackage{booktabs}
\usepackage{natbib}
\usepackage{bbm}
\usepackage{algorithm}
\usepackage[noend]{algpseudocode}
\usepackage{multirow}
\algdef{SE}[DOWHILE]{Do}{doWhile}{\algorithmicdo}[1]{\algorithmicwhile\ #1}
\usepackage[letterpaper, portrait, margin=1in]{geometry}
\usepackage{booktabs}
\usepackage{tabularx}
\usepackage{xcolor}
\hypersetup{
    colorlinks,
    linkcolor={red!50!red},
    citecolor={blue!50!blue},
    urlcolor={blue!80!blue}
}
\usepackage{optidef}

\usepackage{amsthm}

\theoremstyle{definition}

\usepackage{newtxtext} 
\DeclareMathOperator*{\argmax}{arg\,max}

\makeatletter
\newcommand*{\centerfloat}{%
  \parindent \z@
  \leftskip \z@ \@plus 1fil \@minus \marginparwidth
  \rightskip \leftskip
  \parfillskip \z@skip}
\makeatother

\usepackage{amssymb,amsmath,caption}
\usepackage[hang,flushmargin]{footmisc}
\newcommand{\algorithmfootnote}[2][\footnotesize]{
  \let\old@algocf@finish\@algocf@finish
  \def\@algocf@finish{\old@algocf@finish
    \leavevmode\rlap{\begin{minipage}{\linewidth}
    #1#2
    \end{minipage}}%
  }%
}

\usepackage{setspace}
\usepackage{natbib}
\usepackage{algorithm}
\usepackage[noend]{algpseudocode}
\algdef{SE}[DOWHILE]{Do}{doWhile}{\algorithmicdo}[1]{\algorithmicwhile\ #1}

\usepackage{appendix}

\usepackage{array}
\usepackage{threeparttable}
\usepackage{multirow}
\usepackage{rotating}
\usepackage{makecell}

\journal{.}

\biboptions{authoryear}

\begin{document}
\begin{frontmatter}

\title{Traffic Light Control with Reinforcement Learning}


\author[label1]{Taoyu Pan}
\ead{pty060110@gmail.com}
\address[label1]{Cambridge A Level Centre, Hangzhou Foreign Language School, Hangzhou, China, 310000}

\begin{abstract}
Traffic light control is important for reducing congestion in urban mobility systems.  This paper proposes a real-time traffic light control method using deep Q learning. Our approach incorporates a reward function considering queue lengths, delays, travel time, and throughput. The model dynamically decides phase changes based on current traffic conditions. The training of the deep Q network involves an offline stage from pre-generated data with fixed schedules and an online stage using real-time traffic data. A deep Q network structure with a ``phase gate'' component is used to simplify the model's learning task under different phases. A ``memory palace'' mechanism is used to address sample imbalance during the training process. We validate our approach using both synthetic and real-world traffic flow data on a road intersecting in Hangzhou, China. Results demonstrate significant performance improvements of the proposed method in reducing vehicle waiting time (57.1\% to 100\%), queue lengths (40.9\% to 100\%), and total travel time (16.8\% to 68.0\%) compared to traditional fixed signal plans.
\end{abstract}

\end{frontmatter}


\section{Introduction}\label{intro}

With the rapid process of urbanization, car ownership has been constantly increasing, especially in China. According to the statistics of the Ministry of Public Security of China \citep{num_motor2023}, the number of motor vehicles owned nationwide reached 417 million in 2022. The rapid increase in car ownership causes great traffic congestion, resulting in a waste of time for drivers and commuters. 

Traffic light control is crucial for reducing congestion in urban mobility systems. However, traditional traffic control methods typically rely on pre-defined timing schedules and fixed signal patterns to regulate the flow of traffic at intersections \citep{miller1963settings}. While traditional methods have been effective in regulating traffic flow to a certain extent, the limitations of their fixed schedules make them inadequate for addressing the complex and dynamic nature of modern traffic systems. For example, during peak traffic hours, the green light of the major traffic flow direction may be too short to allow enough vehicles to pass through the intersection, causing long queues and delays. Additionally, traditional methods may not take into account the impact of unexpected events such as accidents, road closures, and sudden increase or decrease in traffic flows, which further exacerbate congestion. 

Recent developments in deep reinforcement learning (RL) have shown great promise in improving traffic light control by enabling real-time adaptation to dynamic traffic conditions \citep{li2016traffic}. RL allows the traffic light to learn from its environment by observing traffic patterns and adjusting its control strategy accordingly. The traffic light is trained using a reward-based system, where the goal is to maximize the flow of traffic through the intersection while minimizing delays and reducing congestion. This adaptive approach to traffic control has shown great potential to improve traffic flows and reduce travel time.

This paper proposes a deep Q learning-based RL method for real-time traffic light control. Following the work of \citet{wei2018intellilight}, the proposed method considers a reward function that takes into account queue lengths, delays, travel time, and throughput. At each time step, the model decides whether to change the phase or not, providing an adaptive solution to dynamic traffic conditions. The training process is divided into two stages: offline and online. The offline training uses pre-generated traffic flow data with fixed time schedules to obtain a good prior for model parameters, while the online training leverages real-time traffic flow data for further adaptive learning of the model. To better capture the dynamics of different traffic light phases, a well-designed deep Q network structure with a ``phase gate'' component is employed. Additionally, to address the sample imbalance issue in the experience replay of the deep Q learning, a ``memory palace'' mechanism is used to ensure sufficient sampling of rarely appeared state-action combinations. The proposed approach is validated using both synthetic and real-world traffic flow data, with a road intersection in Hangzhou, China serving as the case study. Results show that our method outperforms traditional fixed signal plan traffic light control in terms of reducing vehicle waiting time (by 57.1\%$\sim$100\%), queue lengths (by 40.9\%$\sim$100\%), and total travel time (by 16.8\%$\sim$68.0\%) in different traffic flow scenarios. The codes and data of this paper are available in \url{https://github.com/OscarTaoyuPan/TrafficLightControl_QS}.

The remainder of the paper is organized as follows. The literature review is shown in Section \ref{sec_liter}. In Section \ref{sec_method}, we describe the methodology including preliminaries, problem definition, and the proposed reinforcement framework. We apply the proposed framework to a Hangzhou road intersection as a case study in Section \ref{sec_case_study}. Finally, we conclude our study and summarize the main findings in Section \ref{sec_con_dis}.

\section{Literature review}\label{sec_liter}
\subsection{Traditional traffic light control}
Traditional traffic light control methods have been widely used for decades and are still prevalent in many cities around the world. These methods are typically based on fixed-time schedules or traffic-responsive strategies that adjust signal timings based on traffic volume or occupancy. Fixed-time schedules use pre-determined signal timings that are set according to historical traffic patterns or the peak traffic volume of a specific area. For example, the fixed-time schedules are set using historical traffic demand to determine the time for each phase \citep{dion2004comparison, miller1963settings, webster1958traffic}. The traffic-responsive strategies use updated time information that is set according to real-world traffic data. For example, \citet{porche1999adaptive} and \citet{cools2013self} implement self-organizing traffic lights using real-time traffic data. These methods can deal with highly random traffic conditions.

One of the main drawbacks of fixed-time schedules is that they cannot adapt to changing traffic conditions, such as fluctuations in traffic volume or unexpected events. This often results in inefficient traffic flow, with long queues and delays and wasted time and fuel consumption. The problem with traffic-responsive strategies is that they are dependent on man-made rules for current traffic patterns, but do not consider the subsequent traffic conditions. In this way, it is unable to generate the optimal solution.

\subsection{Applications of reinforcement learning}
Reinforcement learning (RL) is an area of machine learning that studies how intelligent agents should take actions in an environment in order to maximize the notion of cumulative reward. With the success of Alpha Go \citep{mnih2015human}, RL has gained considerable attention in many fields, including energy \citep{perera2021applications}, civil engineering \citep{fu2022applications}, network system \citep{al2015application}, finance \citep{hambly2023recent}, logistics \citep{yan2022reinforcement}, and transportation \citep{haydari2020deep}. For a brief survey with recent advances in reinforcement learning, people can refer to \citet{arulkumaran2017deep}.

\subsection{Reinforcement learning-based traffic light control}
Since traditional traffic light methods are not able to solve the multi-direction dynamic traffic light control problems in a comprehensive way, there are more and more attempts using RL to deal with the problem \citep{kuyer2008multiagent, mannion2016experimental}. Traditional RL algorithms designed the state as discrete values of traffic conditions, such as the location of vehicles and the number of cars on the road \citep{wiering2000multi, abdulhai2003reinforcement, bakker2010traffic, abdoos2013holonic, el2013multiagent}. However, the corresponding paired action-state matrix has a large space demand for storage. In this way, if larger state space problems are considered, the method will not work.

To solve the problem in order to consider a larger state space, the algorithm of Deep Q learning is applied to take continuous variables into account \citep{li2016traffic}. Deep Q learning sets up a deep neural network (DNN) to learn the Q function of RL from the traffic state inputs and the corresponding traffic system performance output. In this way, the state and action are associated with reward. The state design considers queue length \citep{li2016traffic} and average delay \citep{van2016coordinated}, etc. The reward design also takes those variables into account. Nevertheless, these methods assume relatively static traffic environments, and hence far from real-world traffic conditions. 

Recently, \citet{wei2018intellilight} proposed an RL-based traffic light control model and tested the algorithms in a more realistic traffic setting. In this paper, we follow the framework of \citet{wei2018intellilight}, but using the real-world traffic flow data from a road intersection in Hangzhou, China, which none of the existing studies has considered.

\section{Methodology}\label{sec_method}

\subsection{Preliminaries}
\subsubsection{Introduction to reinforcement learning}

RL is a machine learning paradigm \citep{szepesvari2010algorithms} that aims to control a system by maximizing a long-term objective. RL is based on the Markovian Decision Processes (MDPs) framework \citep{sutton2018reinforcement} which is defined by a set of terms $(\mathcal{S}, \mathcal{P}, \mathcal{A}, R, \gamma)$, where $\mathcal{S}$ is the state space, $\mathcal{P}$ is the probability of transition, $\mathcal{A}$ is the action space, $R$ is the reward, and $\gamma$ is the discount factor.

\textbf{State Space} $\mathcal{S}$: The state space $\mathcal{S}$ is a finite set of Markov states $s_t$ of the environment that can be used by the agent to decide the next action. If a state is Markovian, the history of states can be ignored, and the agent can rely solely on the current state $s_t$ instead of the whole history.

\textbf{Action Space} $\mathcal{A}$: The action space $\mathcal{A}$ is a set of legal actions $a_t$ that can be taken by the agent. At each time step $t$, the agent selects the optimal action from the action space $\mathcal{A}$ following a policy $\pi$ that maximizes the long-term expected return.

\textbf{Transition probability} $\mathcal{P}$: The transition probability $\mathcal{P}$ is defined for each triplet $(s_t, a_t , s_{t+1}) \in \mathcal{S}\times\mathcal{A}\times\mathcal{S}$ and gives the probability of moving from state $s_t$ to state $s_{t+1}$ by taking action $a_t$.

\textbf{Reward} $R$: The reward $R$ is the scalar variable given by a reward function $r$: $\mathcal{S}\times\mathcal{A} \rightarrow R$, obtained by the agent after performing an action $a_t$ at time step $t$ according to its policy. The expected return $G_t$ is defined as the total discounted reward starting from the time step $t$ up to time $t+n$ (where $n$ can be infinity):
\begin{align}
    G_t = R_t + \gamma R_{t+1} + ... + \gamma^n R_{t+n} = \sum_{k=0}^n\gamma^k R_{t+k}
\end{align}
where $\gamma \in [0,1]$ is the discount factor multiplying the future expected rewards. It denotes the importance of future rewards versus immediate rewards. 

The objective of the agent is to determine a policy $\pi$, which is a set of rules for selecting an action based on a state, to maximize the total discounted reward. To achieve this, we employ the Q learning algorithm, which is a type of model-free, value-based, and off-policy RL technique introduced by \citet{watkins1992q}. In Q learning, the Q value function estimates the expected discounted reward of an action given a state (i.e., the goodness of a selected action in a given state). We can define the Q value function as follows:
\begin{align}
Q^{\pi}(s,a) = \mathbb{E}[G_t \mid s, a, \pi]
\end{align}
The optimal action-value function $Q^{*}(s,a)$ is the one that produces the maximum expected return. Therefore, the optimal policy $\pi^{*}$ can be determined by selecting the action $a_t$ that yields the highest Q value for the given state $s$:
\begin{align}
a^{*} = \argmax_{a\in\mathcal{A}} Q^{*}(s, a)
\end{align}
The optimal Q value function is governed by the Bellman equation, and it can be learned through an iterative update process. Specifically, the Q value for the current state-action pair $Q_t(s,a)$ is updated by adding a fraction $\alpha$ of the temporal difference (TD) error $\delta_{t+1}$:
\begin{align}
Q_{t+1}(s,a) = Q_t(s, a) + \alpha \cdot \delta_{t+1}
\end{align}
where $0 < \alpha < 1$ is the learning rate. TD error is defined as the difference between the TD-target $y_t$ and the current Q value $Q_t(s,a)$:
\begin{align}
    \delta_{t+1}(s, a) = y_t - Q_t(s, a)
\end{align}
The TD-target $y_t$ is given by:
\begin{align}
\label{eq_q_learning_yt}
y_t = R_t + \gamma \max_{a_{t+1}\in\mathcal{A}} Q_t( s_{t+1}, a_{t+1})
\end{align}
where $R_t$ is the reward at time $t$, and $\gamma$ is the discount factor.

\subsubsection{Deep reinforcement learning}

The traditional Q learning algorithm \citep{watkins1992q} uses a tabular structure to store the values of state-action pairs, which is challenging to implement in cases of high-dimensional or continuous state spaces. To overcome this limitation, deep neural networks can be used as an alternative to approximate the Q value function. These networks can handle high-dimensional or continuous state spaces, and capture complex state features to approximate the Q values. The standard deep RL algorithm is the Deep Q network (DQN) \citep{mnih2015human}, which is composed of an input layer that takes states (images or vectors) as input, a number of hidden layers to extract and transform features, and an output layer to approximate the Q values of state-action pairs.

During the learning process, DQN uses a neural network with weights $\boldsymbol{\theta}$ and an experience replay memory to store past experiences. At each time step an experience $e_t = (s_t, a_t, R_t, s_{t+1})$ is added to the memory. To update the Q network, a mini-batch of experiences is randomly sampled from the experience replay memory, and the Q learning updates are applied using target values $y_t$ as defined in Eq. \ref{eq_q_learning_yt}:
\begin{align}
\label{eq_dqn_yt}
y_t = R_t + \gamma \max_{a_{t+1} \in \mathcal{A}} Q_t(s_{t+1}, a_{t+1}; \boldsymbol{\theta}_t)
\end{align}
where $\boldsymbol{\theta}_t$ is the weights of the DQN at time step $t$. The learning process aims to have a Q network that accurately predicts the target values in Eq. \ref{eq_dqn_yt}. Thus, the objective of the learning algorithm is to minimize the loss function:
\begin{align}
\label{eq_loss_function}
Loss_t(\boldsymbol{\theta}) = \mathbb{E}\Big[\big(y_t - Q_t(s_t, a_t; \boldsymbol{\theta}_t)\big)^2\Big]
\end{align}

\subsection{Problem definition}
\begin{figure}[htb]
    \centering
    \includegraphics[width = 0.65 \linewidth]{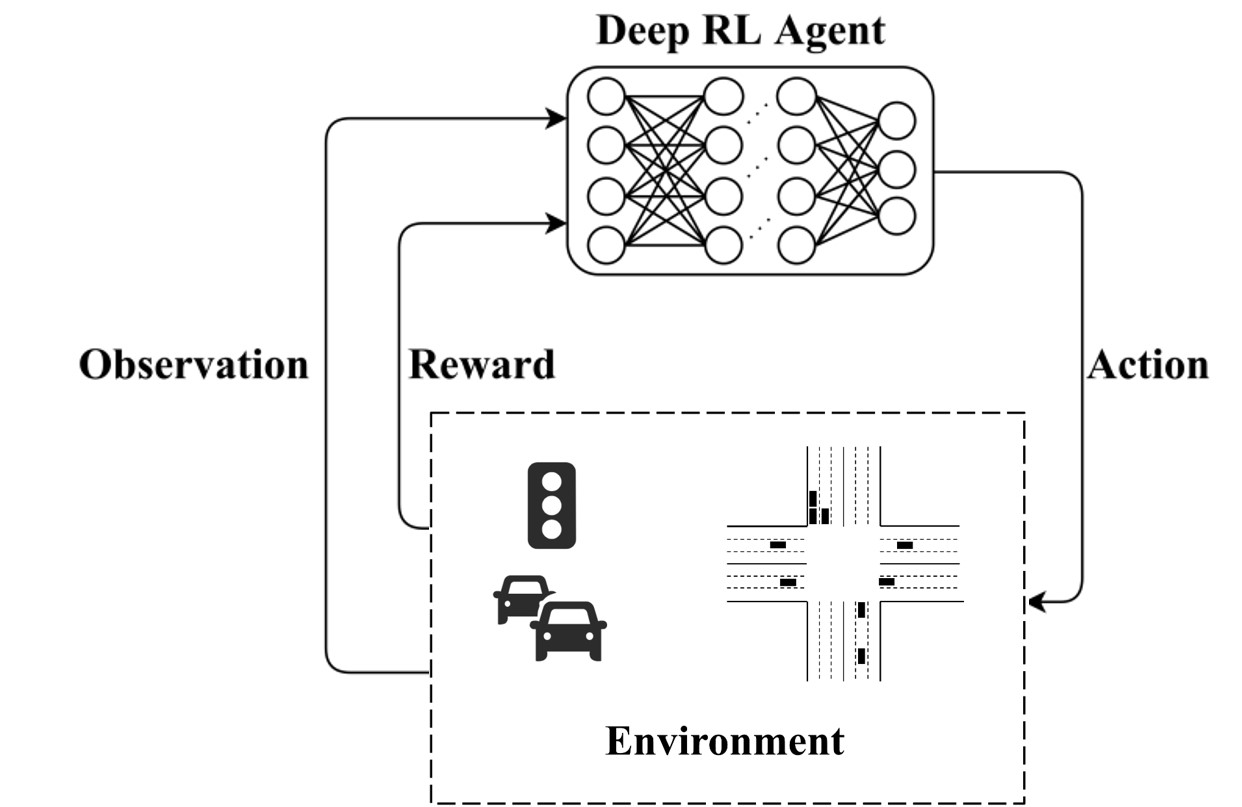}
    \caption{Problem definition}
    \label{fig_prob}
\end{figure}

Figure \ref{fig_prob} provides an overview of the problem addressed in this study. The environment is a traffic signal intersection, and the deep RL agent receives a state $s_t \in \mathcal{S}$, selects an action $a_t \in \mathcal{A}$, and receives a reward $R_t$ from the environment. The goal of the agent is to determine the optimal action $a_t$ for each state $s_t$, with the aim of minimizing the average pre-defined cumulative discounted return. To maintain effective decision-making over time in the dynamic traffic signal intersection environment, the agent must continuously learn and improve its policy. We elaborate on each of the key components of the problem definition in the following.

\textbf{Phase}. In this study, a traffic signal's phase is defined as a specification of signal color for each direction. Two phases are considered: NS and WE, where NS represents green for north and south directions and red for east and west directions and WE represents green for east and west directions and red for north and south directions. Yellow lights are ignored in this study as they have a fixed time length and can be attached to the end of each phase.

\textbf{Environment}. The intersection environment in this study consists of four directions: East (E), West (W), North (N), and South (S). Each direction has a specific lane layout (e.g., three lanes). The environment also includes pre-defined vehicle movements for each phase, such as straight movements for East-West and left turns for East-West in the WE phase.

\textbf{Agent}. The agent plays a crucial role in this study. It observes the state of the environment (traffic signal intersection), selects an action based on its policy, and receives an immediate reward at each time step. Fig. \ref{fig_prob} illustrates the general structure of the interaction between the agent and the environment

\subsection{Model design}
In this section, we describe the specific design for the model, including state, action, and reward functions.
\subsubsection{State description}\label{sec_state}
\begin{figure}[htb]
    \centering
    \includegraphics[width = 0.8 \linewidth]{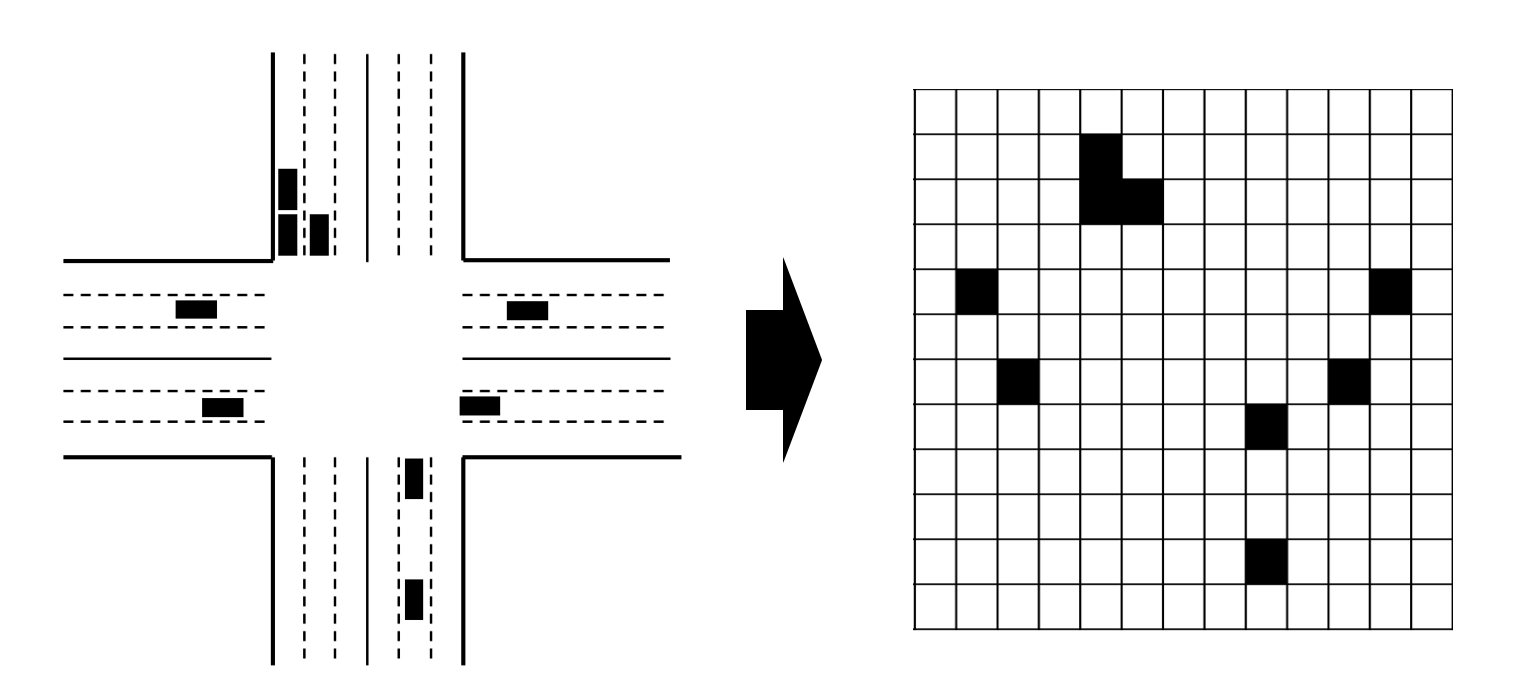}
    \caption{Representation of vehicle's position}
    \label{fig_veh_loc}
\end{figure}

The state is defined as a snapshot of the environment (i.e., intersection). Denote $\mathcal{I}$ as the set of all lanes for four directions. The state at time $t$ includes:
\begin{itemize}
    \item Queue length: ${q_t} = (q_{i,t})_{i\in\mathcal{I}}$, where $q_{i,t}$ indicates the queue length for lane $i$ at time $t$.
    \item Number of vehicles: $v_t = (v_{i,t})_{i\in\mathcal{I}}$, where $v_{i,t}$ is the number of vehicles lane $i$ at time $t$.
    \item Total waiting time: $w_t = (w_{i,t})_{i\in\mathcal{I}}$, where $w_{i,t}$ is the total waiting time (i.e., from the most recent vehicle stop up to time $t$) of all vehicles in lane $i$ at time $t$. 
    \item Phase: ($P_t$, $P_{t+1}$), where $P_t$ is current phase and $P_{t+1}$ is the next phase.
\end{itemize}
Besides the numerical environmental information above, the state includes an image representation of vehicles' positions. As shown in Figure \ref{fig_veh_loc}, vehicle locations at time $t$ are mapped to an image with grids (denoted as $M_t$). In each pixel of the image, value 1 represents there is a vehicle in that grid, otherwise 0. The image is encoded by a convolutional neural network (CNN) to get a latent vector $l_t$:
\begin{align}
l_t = \text{CNN}(M_t;\; \boldsymbol{\theta}_t^{\text{CNN}})
\end{align}
where $\boldsymbol{\theta}_t^{\text{CNN}}$ is the network weights of the CNN at time $t$. 

Therefore, the final state of the model is 
\begin{align}
    \label{eq_state}
    s_t = \text{Concat}(q_t, v_t, w_t, P_t, P_{t+1}, l_t)
\end{align}
where $\text{Concat}(\cdot)$ represents concatenate of all vectors.

\subsubsection{Action set}
There are two actions considered in this study: 1) change to the next phase and 2) keep the current phase. That is, 
\begin{align}
    \mathcal{A} = \{\text{Change to next phase}, \text{Keep the current phase} \}
\end{align}
With this action set, the agent can use current traffic conditions to dynamically determine the cycle length at each time $t$. Note that it does not make sense to switch phases frequently. Therefore, there will be a cost imposed for taking the action of ``change to the next phase''. 

\subsubsection{Reward function}
Reward is the key component of this study. Recall that the objective of the paper is to maximize the flow of traffic through the intersection while minimizing delays and reducing congestion, which is typically used in many transportation-related studies \citep{mo2023individual,mo2023robust, mo2023ex}. To achieve this objective, the reward function at time $t$ is defined as a weighted sum of the various factors. These factors include:
\begin{itemize}
\item Total delays for lane $i$ at time $t$ (denoted as $d_{i,t}$):
\begin{align}
    d_{i,t} = 1- \frac{\text{Avg lane speed}_{i,t}}{\text{Speed limit}_i}
\end{align}
where ``Avg lane speed'' is the mean speed of all vehicles at lane $i$ and time $t$. ``Speed limit'' is the pre-determined lane speed limit.
\item Total waiting time for lane $i$ (denoted as $w_{i,t}$): The waiting time for vehicle $j$ in lane $i$ at time $t$ (denoted as $w_{i,t}^{(j)}$) is defined as the time from its most recent stop (speed $<$ 0.1 m/s) to time $t$. Hence, $w_{i,t}^{(j)}$ will be recounted every time the vehicle moves (i.e., speed $\geq$ 0.1 m/s). Then the total waiting time for lane $i$ is $w_{i,t} = \sum_j w_{i,t}^{(j)}$. 
\item Total queue length of lane $i$ at time $t$ (denoted as $q_{i,t}$): number of vehicles of speed equal to zero at lane $i$ time $t$ 
\item Change of phase (denoted as $C_t$): $C_t = 1$ if the $a_t$ is changing to the next phase. 
\item Total number of vehicles that passed the intersection during time interval $t$ (denoted as $V_t$).
\item Total travel time of all vehicles that passed the intersection during time interval $t$ (denoted as $T_t$). 
\end{itemize}

Given the definition of these factors, the reward function at time $t$ is defined as:
\begin{align}
    R_t = \beta_1 \sum_{i\in\mathcal{I}} d_{i,t} + \beta_2 \sum_{i\in\mathcal{I}} w_{i,t}+ \beta_3 \sum_{i\in\mathcal{I}} q_{i,t} + \beta_4 C_t + \beta_5 V_t + \beta_6 T_t
\end{align}
where $\beta_1,...,\beta_6$ are weights determining the importance of each factor in the reward.

\subsection{Deep Q Network structure}
\begin{figure}[htb]
    \centering
    \includegraphics[width=1\linewidth]{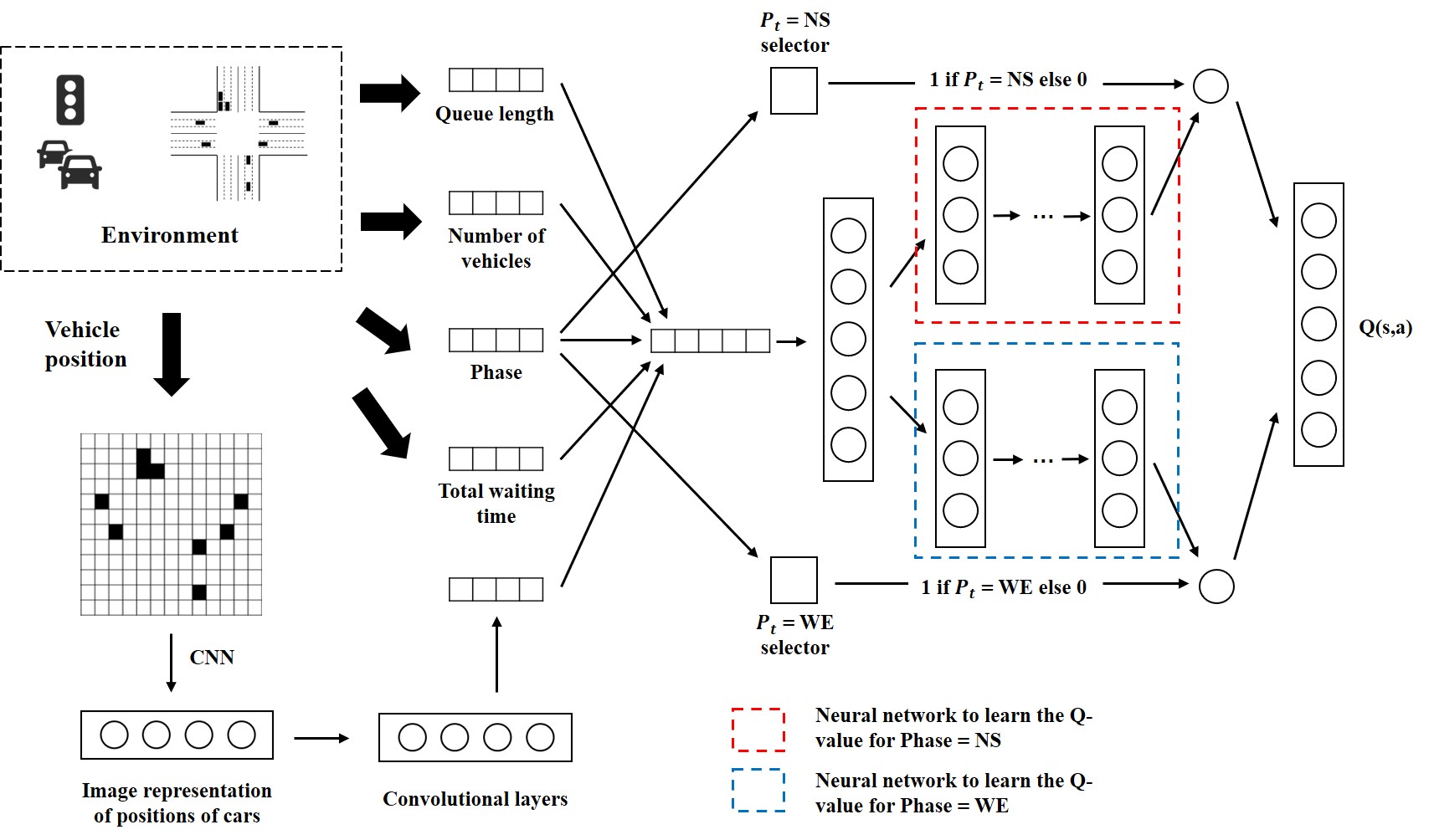}
    \caption{Structure of the deep Q network}
    \label{fig_dqn}
\end{figure}

The structure of the deep Q network is shown in Figure \ref{fig_dqn}. All state information (see Eq. \ref{eq_state} for details) is concatenated as an input vector before being put into the deep Q network. 

In the real-world scenario, traffic conditions can be very different under different phases. For instance, when the system is in phase NS, more traffic on the WE direction should make the light tend to change. However, when the system is in phase WE, the traffic in the WE direction should make the light tend to keep. This means that the traffic on WE directions has two different roles under different phases. The agent should be able to intelligently differentiate this. Simply adding phase information into the state may not be enough. In this study, we implement a deep Q network structure that can consider the different cases explicitly, referred to as ``phase gate''

As depicted in Figure \ref{fig_dqn}, the concatenated features are input to fully connected (FC) layers to learn the mapping from traffic conditions to potential Q values. Separate FC layers (red square and blue square in Figure \ref{fig_dqn}) are employed for each phase to enable a distinct learning process. A phase selector gate is used to determine which branch of the FC layers to activate based on the current phase $P_t$. When $P_t =$ NS, the NS phase selector is set to 1 while the WE phase selector is set to 0, activating the NS branch. Similarly, when $P_t =$ WE, the WE branch is activated. This approach ensures that the decision-making process is tailored to the specific phase, avoids bias towards certain actions, and improves the network's fitting capability \citep{wei2018intellilight}.

\subsection{Algorithm training}
\subsubsection{Offline pretrain and online training}
\begin{figure}[htb]
    \centering
    \includegraphics[width=1\linewidth]{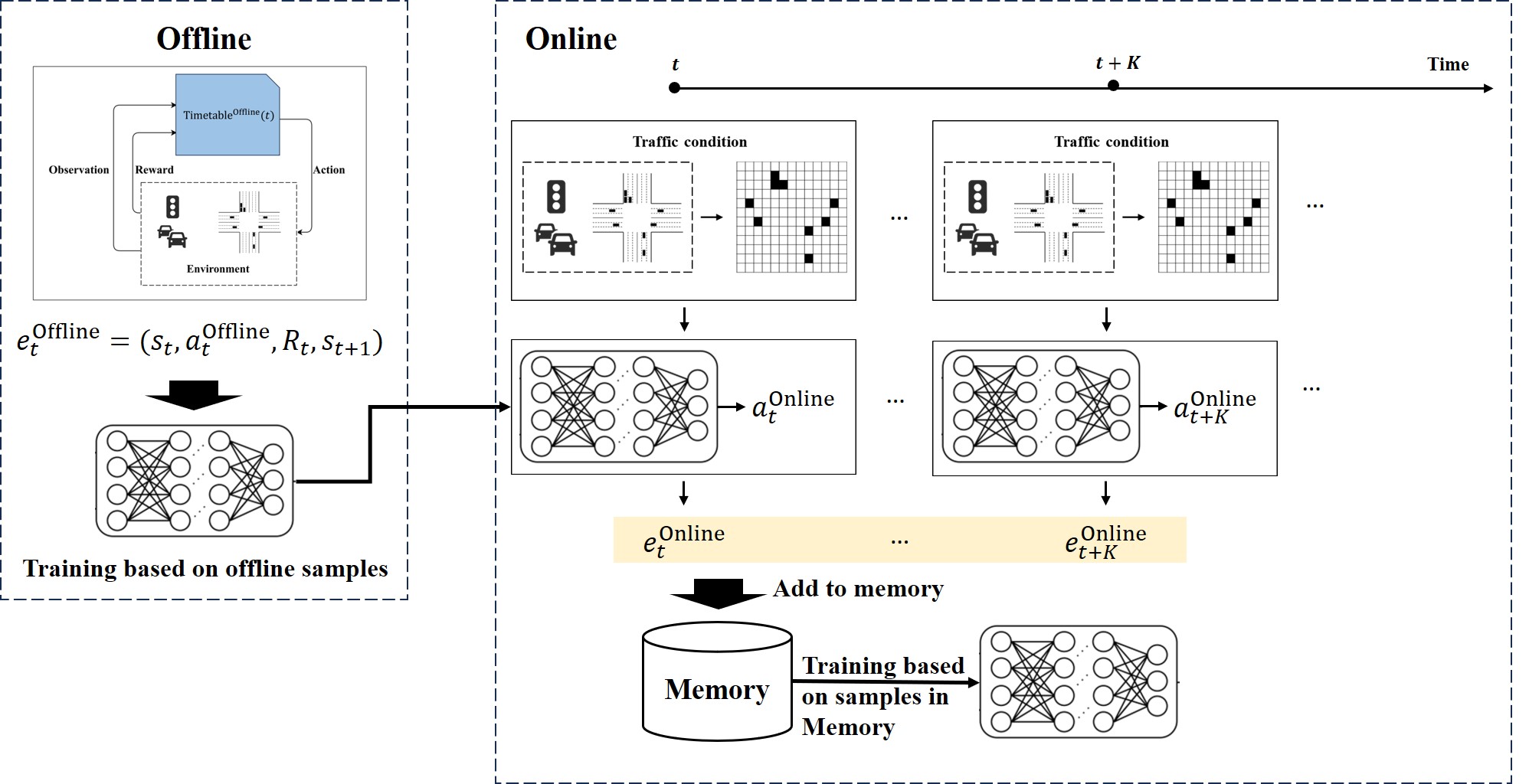}
    \caption{Illusration of offline line pretrain and online training}
    \label{fig_offline_online}
\end{figure}

Our model is composed of two training steps (as shown in Figure \ref{fig_offline_online}): offline and online. In the offline stage, we set several fixed timetables for the lights and let traffic go through the system to collect data samples. Compared to the typical online deep Q learning, the offline training part decides the action based on the pre-determined fixed timetable:
\begin{align}
    a_t^{\text{Offline}} = \text{Timetable}^{\text{Offline}}(t)
\end{align}
where $\text{Timetable}^{\text{Offline}}(\cdot)$ is a function that returns changing the phase or not according to the current time step $t$. For example, if the given fixed timetable is phase NS = 20 seconds, phase WE = 10 seconds, and $t=0$ is phase WE. Then we have $a_t^{\text{Offline}}$ equals to ``Change to the next phase'' for all $t = 10, 30, 40, 60, 70, 90, ...$ (i.e., every time point when we switch the phase according to the 20/10 timetable). And otherwise $a_t^{\text{Offline}}=$``Keep the current phase''.      

At each step $t$ of the offline stage, the collected sample (experience) is $e_t^{\text{Offline}} = (s_t,a_t^{\text{Offline}},R_t,s_{t+1})$. After collecting samples for several timetables, we use all collected samples to pre-train the DQN using the same loss function in Eq. \ref{eq_loss_function}.

After offline training, the pre-trained model is deployed in the online stage. At each time step $t$, the traffic light agent observes the state $s_t$ from the environment and selects an action $a_t^{\text{Online}}$ (i.e., whether to change the traffic signal to the next phase or not) using an $\epsilon$-greedy strategy (Eq. \ref{eq_at_online}) that combines exploration (i.e., taking a random action with probability $\epsilon$) and exploitation (i.e., selecting the action with the highest estimated reward). This strategy allows the agent to balance between exploring new actions and exploiting the learned knowledge to make optimal decisions in real-time traffic scenarios.
\begin{equation}
   a_t^{\text{Online}} = \left\{
\begin{aligned}
 \text{Randomly pick an action from $\mathcal{A}$}  &\quad\text{     with probability $\epsilon$} \\
 \argmax_{a\in\mathcal{A}} Q_t(s_t, a,\;\boldsymbol{\theta}_t) & \quad\text{     otherwise},
\end{aligned}
\right.  \label{eq_at_online}
\end{equation}
After taking action $a_t^{\text{Online}}$, the agent will observe the environment and get the reward $R_t$ from it. Then, the tuple $e_t^{\text{Online}} = (s_t,a_t^{\text{Online}},R_t,s_{t+1})$ will be stored into memory. After $K$ timestamps (i.e., newly collected samples are $(e_t^{\text{Online}}, ..., e_{t+K}^{\text{Online}})$), the agent will update the network according to samples in the memory.

\subsubsection{Phase-action dependent replay memory and Balanced sampling}
In deep Q learning, the agent collects samples at every time step and then uses these samples to update the deep Q network. Typically, a memory buffer is used to store all these samples. New samples are added while old samples are removed to maintain a constant sample size. This technique is known as experience replay \citep{mnih2015human} that is widely used in RL models. However, in real-world traffic conditions, traffic patterns across different directions can be highly imbalanced. Previous studies \citep{gao2017adaptive,genders2016using,li2016traffic,van2016coordinated} stored all the state-action-reward training samples in a single memory buffer. This memory buffer can be dominated by samples from frequently occurring phases and actions in imbalanced settings. For example, if a road intersection experiences mostly North-South (NS) traffic flows and few East-West (WE) flows, the memory buffer will be dominated by samples with $a_t =$ ``Keep the phase'' and $P_t =$ NS (along with associated $s_t$ and $R_t$), while ignoring less frequent phase-action combinations such as $a_t =$ "Change the phase" and $P_t =$ WE. As a result, the Q values learned for these less frequent phase-action combinations may be inaccurate, leading to poor decision-making by the learned agent for infrequent phase-action combinations.

\begin{figure}[H]
    \centering
    \includegraphics[width=0.65\linewidth]{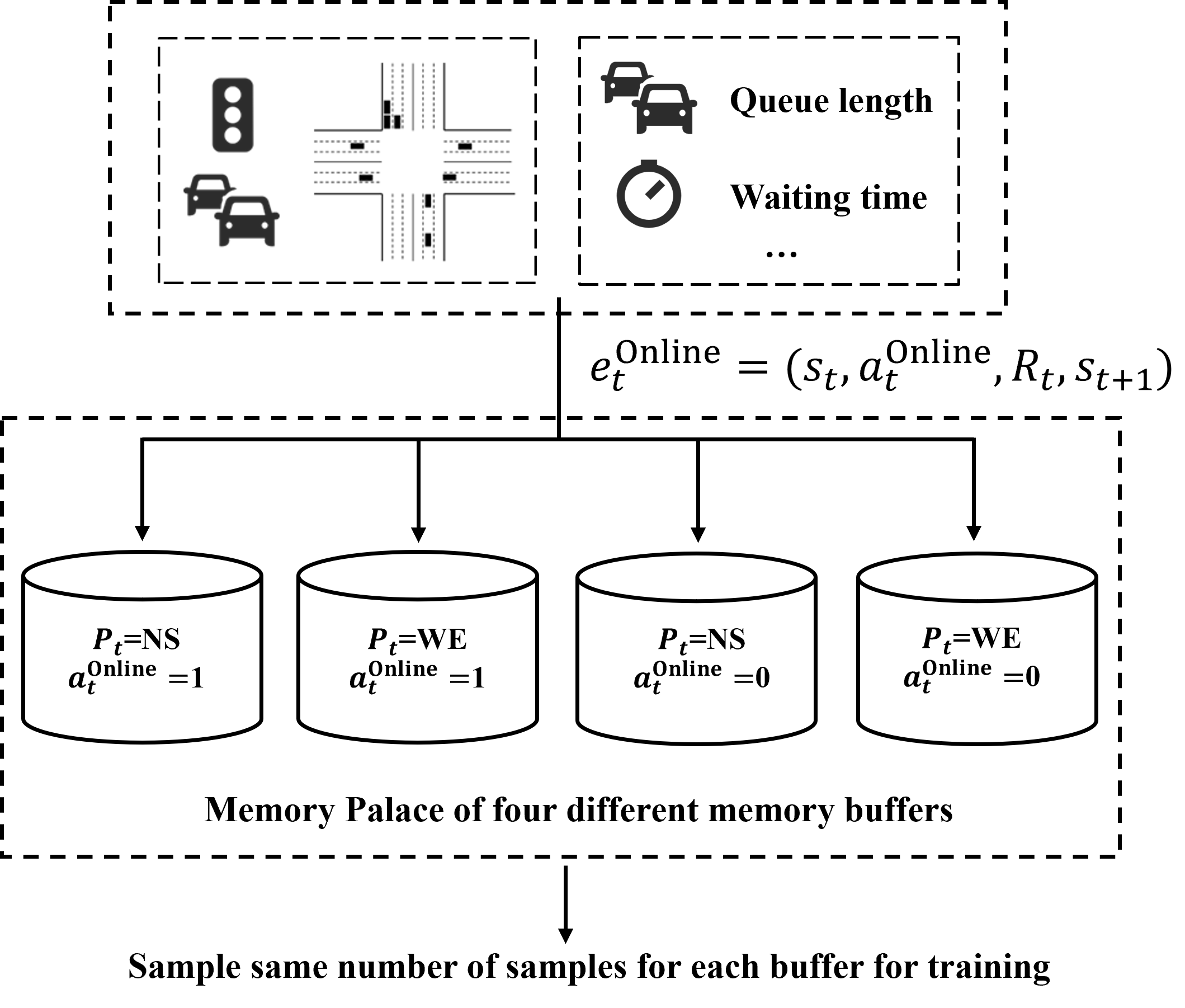}
    \caption{Illustration of memory palace}
    \label{fig_memory}
\end{figure}

To address this issue, this study employs a Memory Palace mechanism \citep{wei2018intellilight}. Instead of using a single memory buffer for all samples as in typical Deep Q learning, we define separate memory buffers for different phase-action combinations, as illustrated in Figure \ref{fig_memory}. Training samples for different phase-action combinations are stored in their respective memory buffers. During each training step, an equal number of samples are selected from different memory buffers to ensure balanced training samples for learning the Deep Q network. This approach prevents interference among different phase-action combinations during the training process, improving the network's ability to accurately predict Q values for each phase-action combination.

\subsection{Simulation tool}
\begin{figure}[htb]
    \centering
    \includegraphics[width = 0.75 \linewidth]{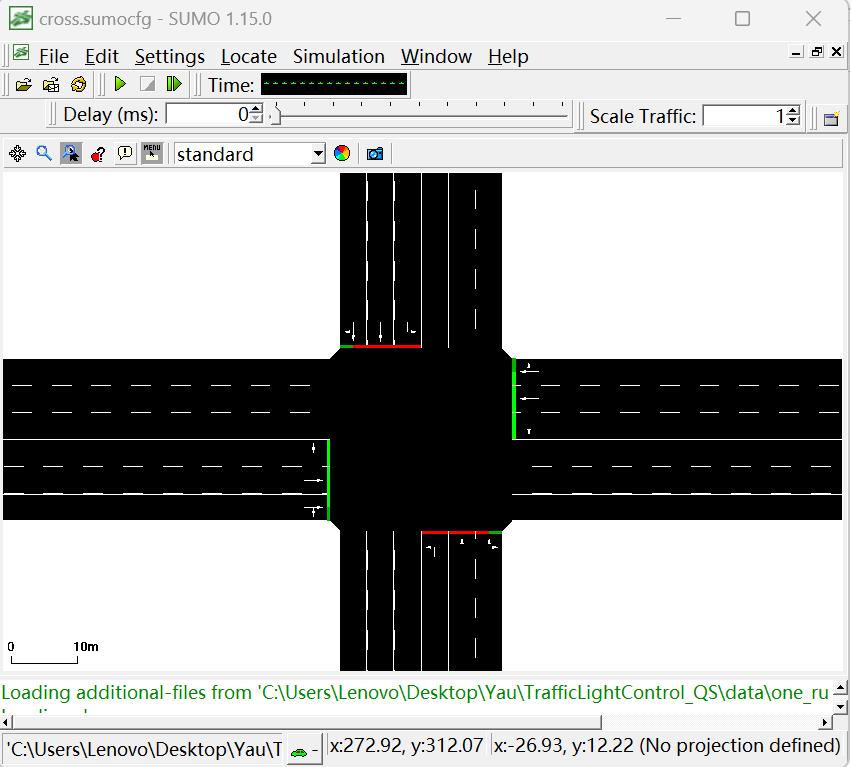}
    \caption{Example of SUMO simulator}
    \label{fig_sumo}
\end{figure}

The SUMO (Simulation of Urban Mobility) tool \citep{SUMO2018} was used to simulate traffic in this study. SUMO is a widely recognized open-source traffic simulator that offers useful application programming interfaces (APIs) and a graphical user interface (GUI) for modeling large road networks and handling them effectively (see Figure \ref{fig_sumo}). It supports dynamic routing based on the right-side driving rules of road intersections and provides a visual graphical interface for designing various road network layouts in multiple grid formats. Additionally, SUMO allows for controlling the traffic lights at each intersection according to user-defined policies. It also enables capturing snapshots of each simulation step, allowing us to obtain the state information $s_t$ for our study. 

\section{Case study design}\label{sec_case_study}

\subsection{Traffic intersection}
\begin{figure}[H]
    \centering
    \includegraphics[width=0.75\linewidth]{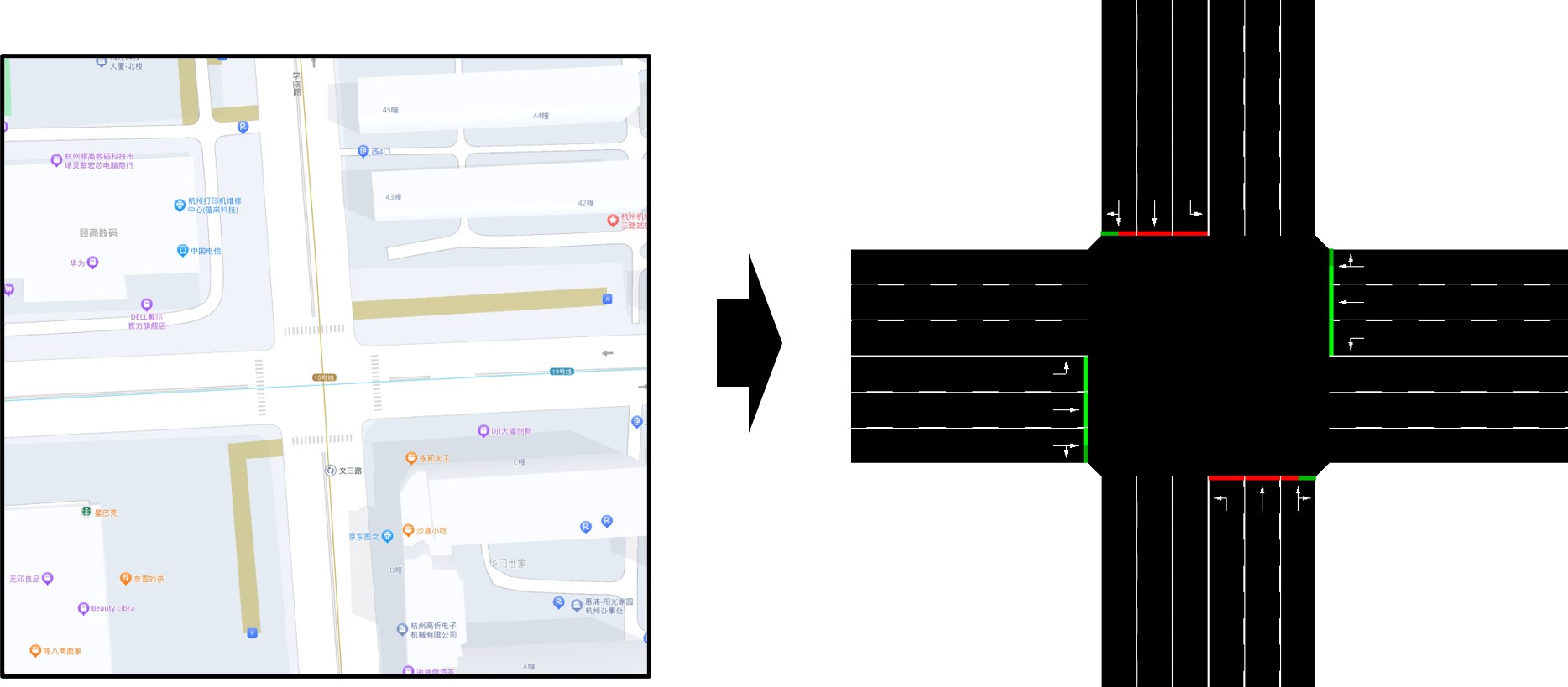}
    \caption{Layout of the case study intersection}
    \label{fig_road_intersection}
\end{figure}

This paper presents a case study of a real-world traffic intersection located at Xueyuan Road and Wensan Road in Hangzhou City. The layout of the intersection is depicted in Figure \ref{fig_road_intersection}. The intersection features four directions, each with three lanes. The rightmost lane is designated for right turns and going straight, the middle lane is exclusively for going straight, and the leftmost lane is reserved for left turns.

\subsection{Experiment design}
To evaluate the effectiveness of the proposed algorithm, we conducted experiments on four distinct traffic conditions as outlined in Table \ref{tab_exp}. The first three scenarios were synthetic, while the last one was based on actual traffic flow data collected from surveillance cameras in Hangzhou. The first scenario represented a balanced traffic situation, where both NS and WE directions had equal vehicle arrival rates of 720 vehicles per hour. The second scenario simulated an imbalanced traffic situation with significantly higher flow rates in the WE direction compared to the NS direction. The third scenario represented an extreme switching situation, where traffic flows were present in either the WE or NE direction for different halves of the simulation time. These synthetic scenarios were designed to assess the model's performance under varying and complex traffic conditions. The last scenario utilized actual flow rate data in Hangzhou from previous studies \citep{zheng2019frap, wei2019colight, wei2019survey}, which originally only covered the morning peak hour from 7:00 AM to 8:00 AM. To allow for longer training time, we duplicated the data to span a 20-hour simulation period.

\begin{table}[htb]
\centering
\caption{Experiment design}\label{tab_exp}
\begin{tabular}{@{}ccccc@{}}
\toprule
\textbf{Scenario}                  & \textbf{Directions} & \textbf{Arrival rate (\# veh/hr)} & \textbf{Start time (hr)} & \textbf{End time (hr)} \\ \midrule
\multirow{2}{*}{Balanced}          & WE                  & 720                               & 0                        & 20                     \\
                                   & NS                  & 720                               & 0                        & 20                     \\\hline 
\multirow{2}{*}{Imbalanced}        & WE                  & 1440                              & 0                        & 20                     \\
                                   & NS                  & 240                               & 0                        & 20                     \\\hline 
\multirow{2}{*}{Switch} & WE                  & 1440                              & 0                        & 10                     \\
                                   & NS                  & 1440                              & 10                       & 20                     \\\hline 
\multirow{2}{*}{Hangzhou}          & WE                  & 716                               & 0                        & 20                     \\
                                   & NS                  & 1132                              & 0                        & 20                     \\ \bottomrule
\end{tabular}
\end{table}

\subsection{Parameter setting}
The parameter settings in this study are similar to \citet{wei2018intellilight}. The time interval between two consecutive time steps ($t$ and $t+1$) is set as 5 seconds. The model update interval is 300 seconds. The discount factor $\gamma$ for future reward is set as 0.8. $\epsilon=0.05$ is used for $\epsilon$-greedy exploration. The batch size for each training is 300. The memory length for each phase-action-based replay buffer is 1000. As the total experiment time is 20 hours, the first 2 hours are used for offline training. The coefficients for the reward function are summarized in Table \ref{tab_reward_para}. 

\begin{table}[H]
\centering
\caption{Coefficients for the reward function}\label{tab_reward_para}
\begin{tabular}{@{}cccccc@{}}
\toprule
$\beta_1$ & $\beta_2$ & $\beta_3$ & $\beta_4$ & $\beta_5$ & $\beta_6$ \\ \midrule
-0.25  & -0.25    & -0.25   & -5   &             -1               &      -1                        \\ \bottomrule
\end{tabular}
\end{table}

\subsection{Benchmark methods}
For comparison, we use the well-known Webster’s Equation \citep{signal2023} to calculate fixed traffic signal plans for each of the four scenarios in Table \ref{tab_exp}. The green phase time for each direction is divided by proportional to the traffic flow volume. The resulting fixed signal plans are shown in Table \ref{tab_fix_signal}

\begin{table}[htb]
\centering
\caption{Fixed signal plans for different scenarios}\label{tab_fix_signal}
\begin{tabular}{@{}cccc@{}}
\toprule
\textbf{Secnario} & \textbf{WE (second)} & \textbf{NS (second)} & \textbf{Cycle length (second)} \\ \midrule
Balanced          & 18                   & 18                   & 36                             \\
Imbalanced        & 33                   & 6                    & 39                             \\
Switch            & 28                   & 28                   & 56                             \\
Hangzhou          & 16                   & 25                   & 41                             \\ \bottomrule
\end{tabular}
\end{table}

\subsection{Results}
\begin{figure}[H]
\centering
\subfloat[Balanced]{\includegraphics[width=0.5\linewidth]{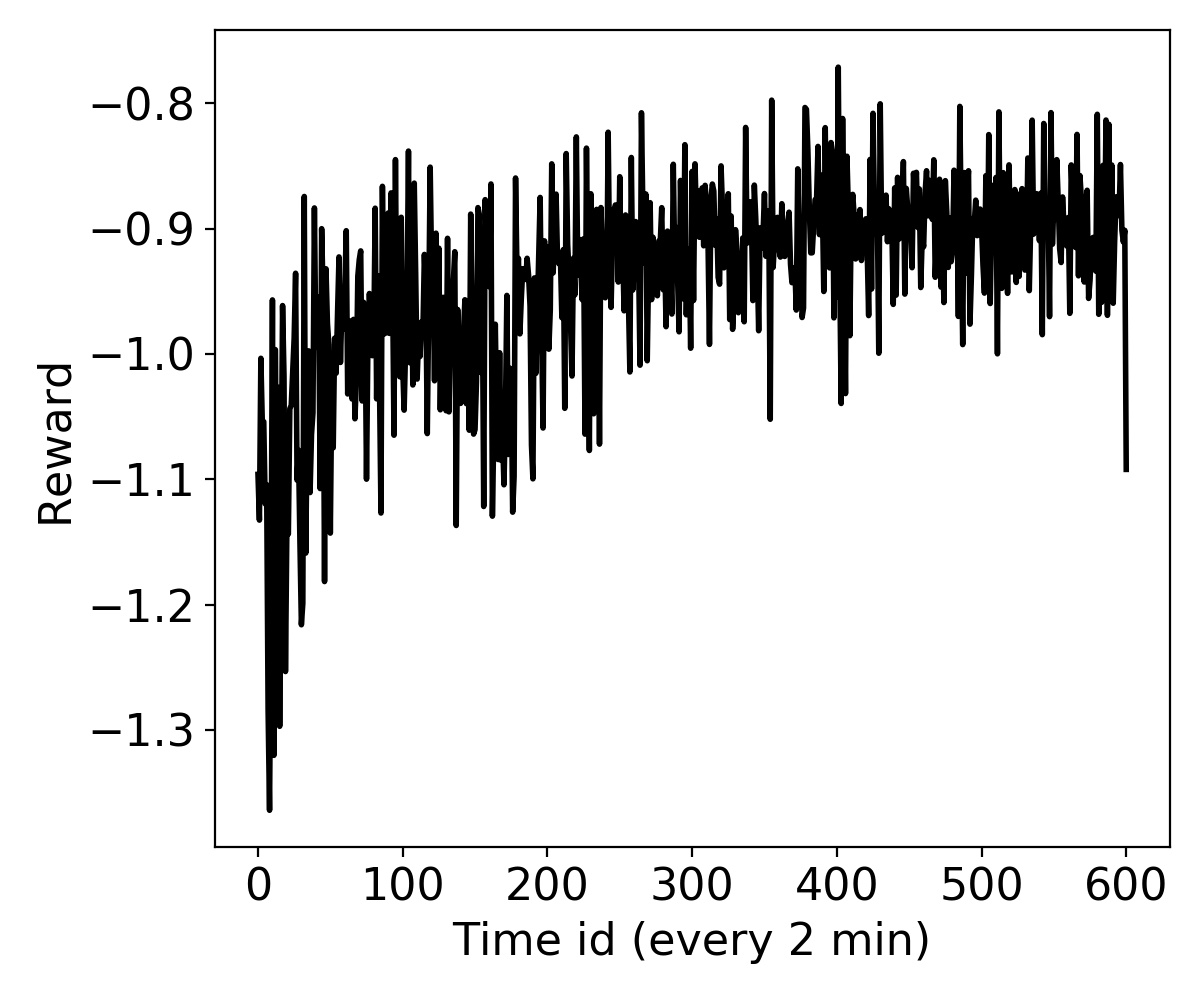}}
\subfloat[Imbalanced]{\includegraphics[width=0.5\linewidth]{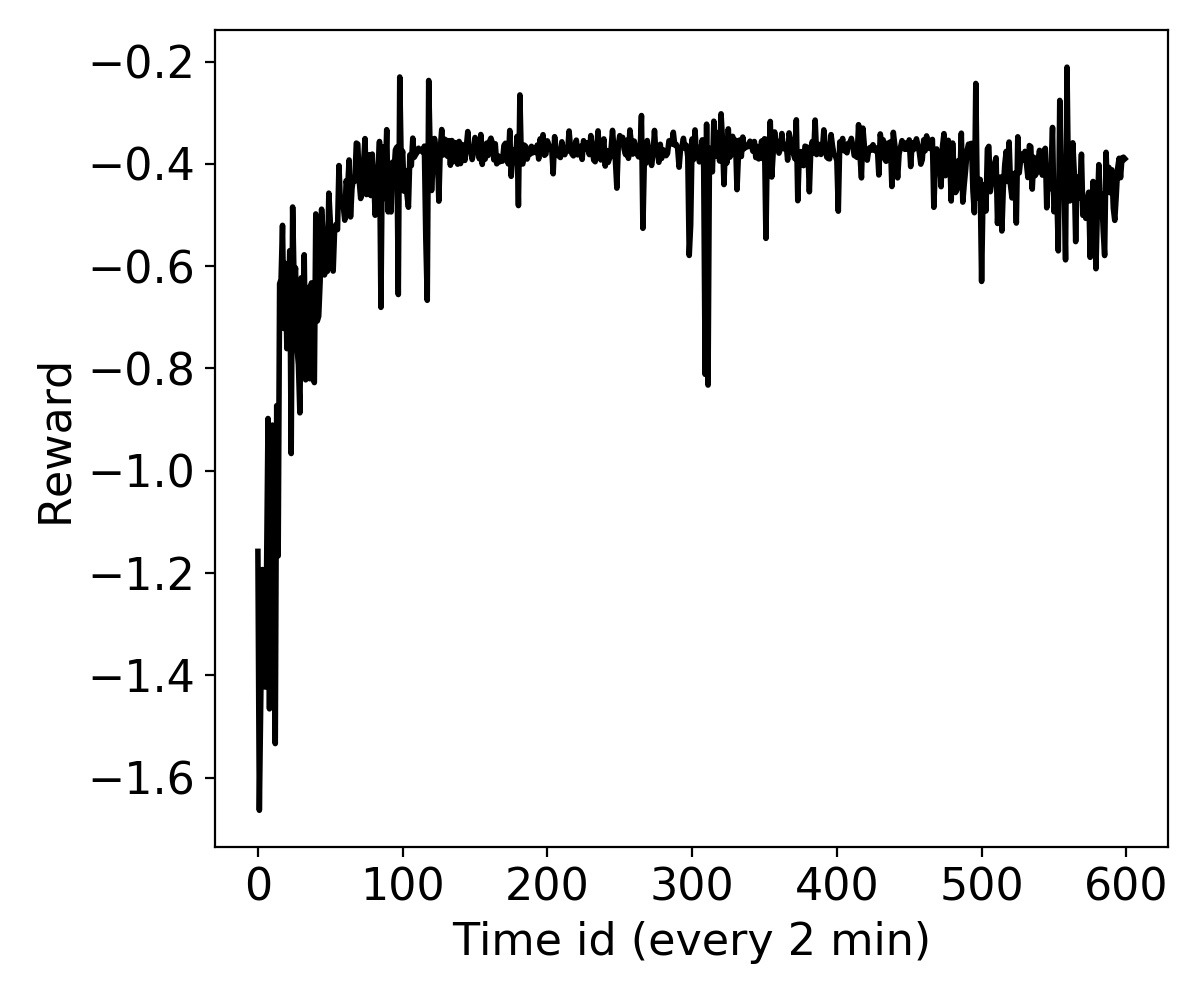}}\\
\subfloat[Switch]{\includegraphics[width=0.5\linewidth]{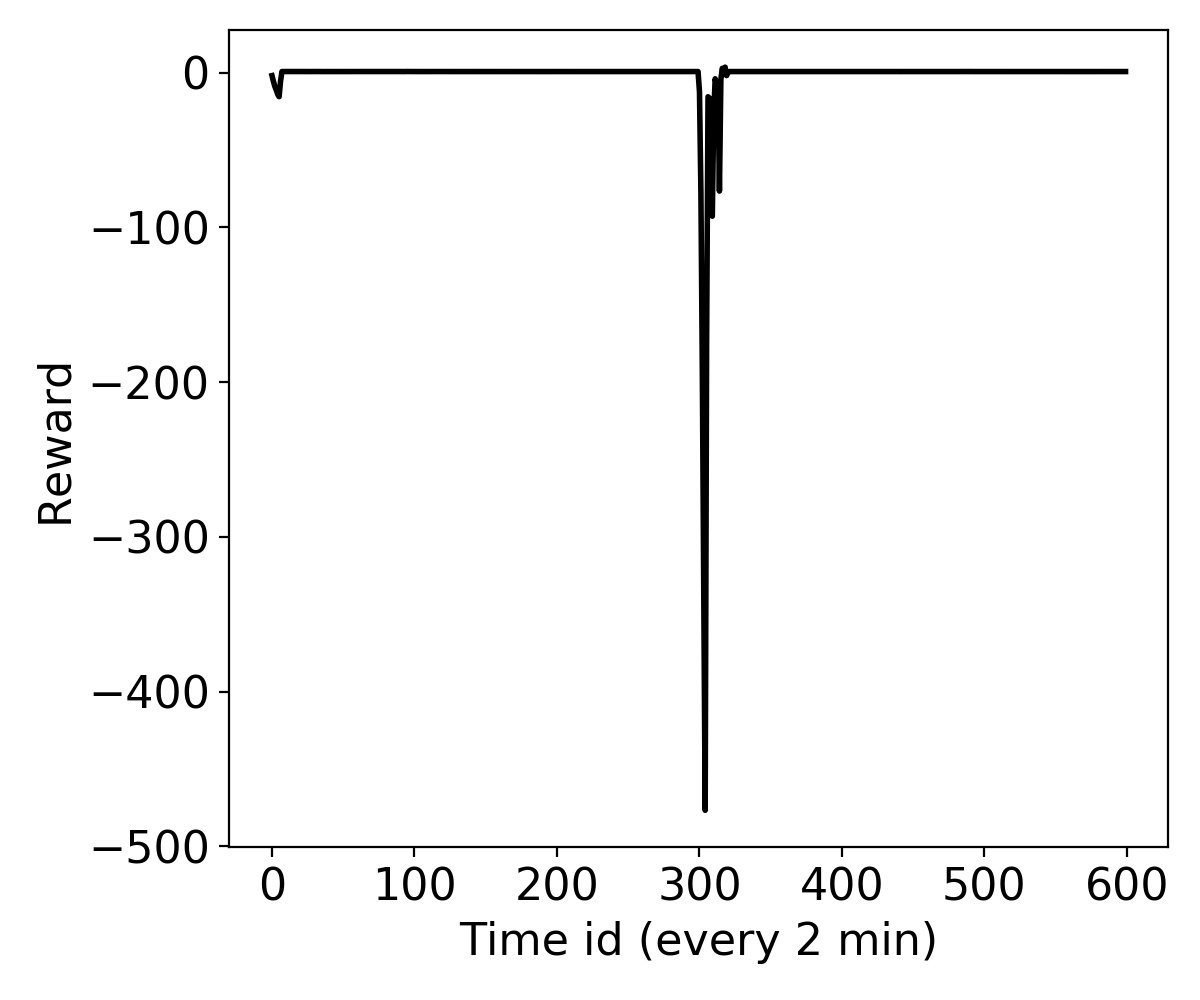}} 
\subfloat[Hangzhou]{\includegraphics[width=0.5\linewidth]{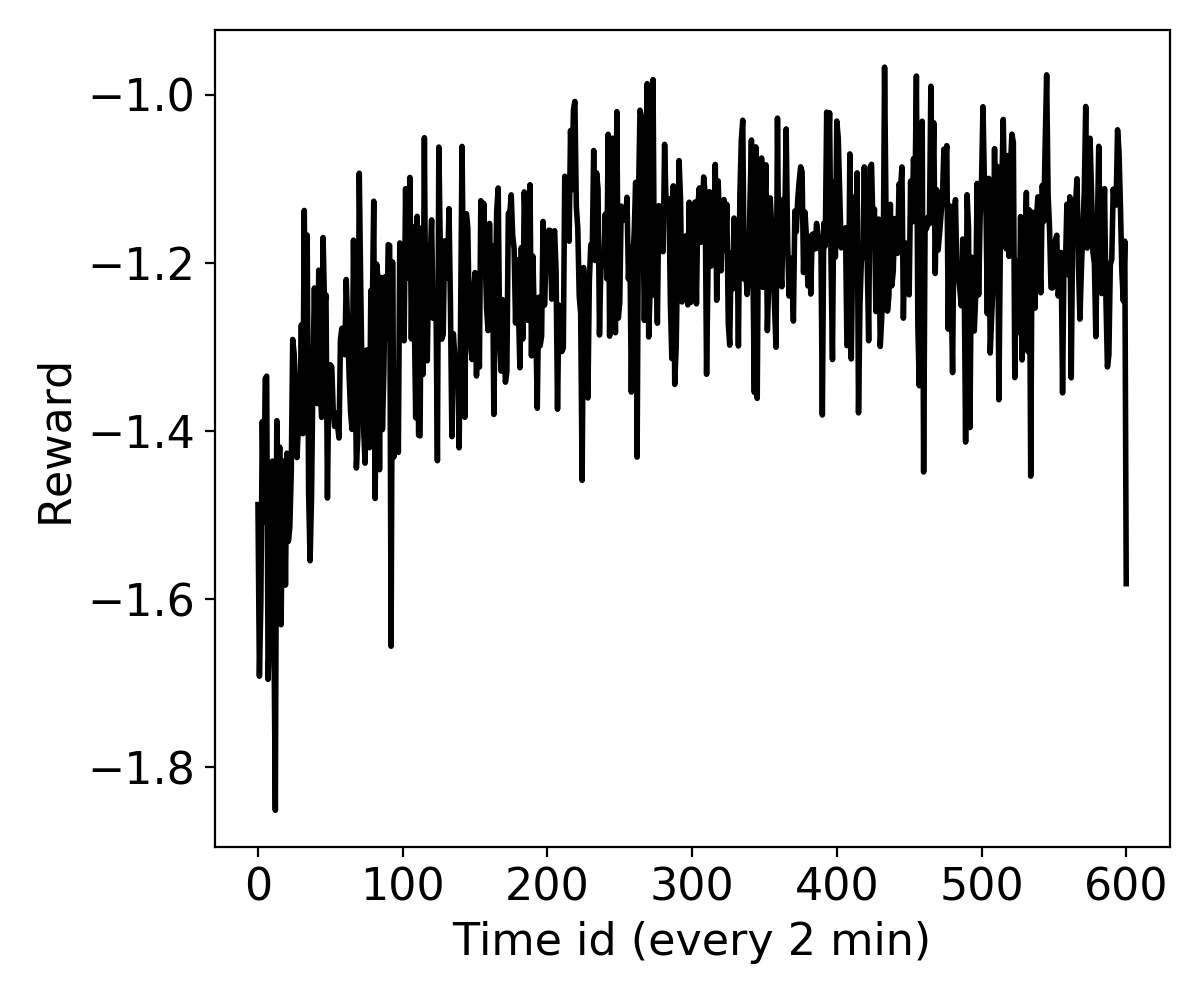}}
\caption{Reward function during the training process}
\label{fig_reward_fun}
\end{figure}

Figure \ref{fig_reward_fun} shows the reward function change during the training process of RL models for four different scenarios. Similar to previous DQN work \citep{mnih2015human}, the training process has fluctuations but the overall reward function keeps increasing, showing that the algorithm keeps learning better traffic light control strategies. 
It is worth noting that in the switching scenario, where one-directional traffic flows change between the first and second half of the simulation period, the strategy is relatively straightforward to learn (i.e., always green for the current direction). As a result, the reward function remains largely unchanged for the majority of the simulation time. However, at the switching point, when the old strategy needs to be reversed, the system experiences a significant drop in the reward function. Fortunately, the RL method quickly adapts and adjusts its strategies to accommodate the dynamic traffic conditions in real time (i.e., changes the direction of the green light). This is evident from the prompt recovery of the reward function. These results demonstrate the capability of the RL approach to dynamically adapt its control strategies in response to changing traffic patterns, ensuring efficient traffic flow management.

The final results of all models and scenarios are summarized in Table \ref{tab_result}, with four metrics selected for comparison: average waiting time of all vehicles, average travel time of all vehicles, average queue length of all lanes, and the reward. These metrics are aggregated over a 1-hour interval. The RL-based method consistently outperforms the fixed signal method across all conditions and metrics, with notable improvements observed, particularly in imbalanced and switch scenarios. This highlights the limitations of traditional static methods in effectively addressing unconventional traffic conditions. It is worth noting that the 0 wait time and queue length for ``switch'' traffic conditions are due to its one-directional traffic flows during a period of time, where the RL model is able to learn that and sets the corresponding traffic signal to be green for the directions with flows. The superiority of the RL approach demonstrates its ability to adapt and optimize traffic control strategies, leading to reduced waiting times, improved travel times, shorter queue lengths, and overall enhanced performance.

\begin{table}[htbp]
\centering
\caption{Model results}\label{tab_result}
\begin{tabular}{@{}cccccc@{}}
\toprule
\multirow{2}{*}{Models} & \multirow{2}{*}{Scenario} & \multicolumn{4}{c}{Performance metrics}   \\ \cmidrule(l){3-6} 
                                &                        & Wait time (sec)            & Travel time (sec)          & Queue length & Reward  \\ \midrule
\multirow{4}{*}{Fixed signal}       
                                & Balanced              & 14.5	& 204.2	& 1.988	& -1.139      \\
                                & Imbalanced            & 13.8	& 200.5	& 1.383	& -0.822      \\
                                & Switch                & 80.3	& 697.2	& 7.717	& -1.691      \\
                                & Hangzhou              & 22.6	& 270.9	& 2.720 & -1.488      
\\\midrule
\multirow{4}{*}{RL}       
                                & Balanced              &6.2 (-57.2\%)	& 169.3 (-17.1\%)	& 1.180 (-40.6\%)	& -0.903 (+20.7\%)      \\
                                & Imbalanced            &1.5 (-89.1\%)	& 162.6 (-18.9\%)	& 0.472 (-65.9\%)	& -0.426  (+48.2\%)      \\
                                & Switch                &0.0 (-100\%)	&222.9 (-68.0\%)	& 0.000	(-100\%)    &0.721 (+142.6\%)       \\
                                & Hangzhou              &9.7 (-57.1\%)	& 226.2 (-16.5\%)	& 1.697 (-37.6\%)	& -1.164 (+21.8\%)
\\ \bottomrule
\multicolumn{6}{l}{
\begin{tabular}[c]{@{}l@{}} The numbers in the parentheses represent percentage change compared to the fixed signal method\\
\end{tabular}} 
\end{tabular}
\end{table}

\section{Conclusion and Discussion}\label{sec_con_dis}
In this paper, we present an RL approach, specifically deep Q learning, to tackle the challenging problem of traffic light control. Our proposed method incorporates a comprehensive reward function that considers queue lengths, delays, travel time, and throughput, enabling an adaptive solution to varying traffic conditions. At each time step, the model intelligently determines whether to change the traffic light phase, allowing for a dynamic response to the evolving traffic environment.

The training process consists of two stages: offline and online. During offline training, we utilize pre-generated traffic flow data with fixed time schedules to establish a strong initial set of model parameters. This offline training phase provides a solid foundation for subsequent adaptive learning during the online training stage, where real-time traffic flow data is leveraged to continually refine the model's performance.

To effectively capture the dynamics associated with different traffic light phases, we employ a well-designed deep Q network structure featuring a unique "phase gate" component. This component ensures that the model focuses on the appropriate information for each specific phase, enhancing its decision-making capabilities. Furthermore, to address the issue of sample imbalance in the experience replay process of deep Q learning, we introduce a novel ``memory palace'' mechanism. This mechanism guarantees sufficient sampling of rarely encountered state-action combinations, thus improving the model's ability to accurately estimate Q values for all possible phase-action pairs.

To validate our approach, we conduct experiments using both synthetic and real-world traffic flow data, with a road intersection in Hangzhou, China serving as our case study. The results demonstrate that our proposed method outperforms traditional fixed signal plan traffic light control in terms of reducing vehicle waiting time (by 57.1\%$\sim$100\%), queue lengths (by 40.9\%$\sim$100\%), and total travel time (by 16.8\%$\sim$68.0\%) in different traffic flow scenarios. Importantly, since our trained model can make real-time decisions, our approach has the potential to be implemented for real-world traffic control scenarios, leveraging up-to-date traffic flow information as input.

Future studies can be pursued in several directions to further advance the field of traffic light control. Two potential avenues for exploration include 1) Extending the model to the multi-intersection case: While this paper focuses on a single intersection, it is crucial to acknowledge that real-world road networks are significantly more complex. Future studies could delve into the interactions between different intersections and explore the application of multiple RL agents for network-level control. By considering the collective behavior of multiple intersections, we can develop more comprehensive and efficient traffic control strategies that optimize the overall network performance. 2) Developing enhanced network structures for information extraction: The current Deep Q network employs a phase gate mechanism, allowing different components to specialize in specific conditions. However, future research can investigate the integration of multiple phases or conditions to further streamline the learning process. For instance, incorporating distinct network components dedicated to peak and off-peak traffic conditions could facilitate more accurate and efficient decision-making. By tailoring the network structure to specific traffic scenarios, we can enhance the model's adaptability and performance in varying traffic conditions. 3) Comparing the proposed approach with more benchmark models, such as simulation-based optimization method \citep{osorio2015urban, mo2021calibrating, mo2020capacity} and other state-or-the-art RL methods.

\bibliography{mybibfile}

\end{document}